\documentclass[sigconf,authorversion]{acmart}
\usepackage{arydshln}
\begin{document}

%
\title[One-shot Image-to-Image Translation]{One-Shot Image-to-Image Translation via Part-Global Learning with a Multi-adversarial Framework}
\author{Ziqiang Zheng}
\affiliation{%
  \institution{Ocean University of China}
  \streetaddress{Songling 238 road}
  \city{Qingdao}
  \state{Shandong}
  \country{China}}
  
  \author{Zhibin Yu}
\affiliation{%
  \institution{Ocean University of China}
  \streetaddress{Songling 238 road}
  \city{Qingdao}
  \state{Shandong}
  \country{China}}
  
    \author{Haiyong Zheng}
\affiliation{%
  \institution{Ocean University of China}
  \streetaddress{Songling 238 road}
  \city{Qingdao}
  \state{Shandong}
  \country{China}}

      \author{Yang Yang}
\affiliation{%
  \institution{University of Electronic Science and Technology of China}
  \streetaddress{No.2006, Xiyuan Ave, West Hi-Tech Zone}
  \city{Chengdu}
  \state{Sichuan}
  \country{China}}
  
        \author{Heng Tao Shen}
\affiliation{%
  \institution{University of Electronic Science and Technology of China}
  \streetaddress{No.2006, Xiyuan Ave, West Hi-Tech Zone}
  \city{Chengdu}
  \state{Sichuan}
  \country{China}}

%

%
%
\begin{abstract}
It is well known that humans can learn and recognize objects effectively from several limited image samples. However, learning from just a few images is still a tremendous challenge for existing main-stream deep neural networks. Inspired by analogical reasoning in the human mind, a feasible strategy is to ``translate'' the abundant images of a rich source domain to enrich the relevant yet different target domain with insufficient image data. To achieve this goal, we propose a novel, effective multi-adversarial framework (MA) based on part-global learning, which accomplishes one-shot cross-domain image-to-image translation. In specific, we first devise a part-global adversarial training scheme to provide an efficient way for feature extraction and prevent discriminators being over-fitted. Then, a multi-adversarial mechanism is employed to enhance the image-to-image translation ability to unearth the high-level semantic representation. Moreover, a balanced adversarial loss function is presented, which aims to balance the training data and stabilize the training process. Extensive experiments demonstrate that the proposed approach can obtain impressive results on various datasets between two extremely imbalanced image domains and outperform state-of-the-art methods on one-shot image-to-image translation.
Our code will be published with the paper.
\end{abstract}

%
%


%

%
%
\maketitle

\section{Introduction}
\begin{figure}[!ht]
\centering
\includegraphics[width=\columnwidth]{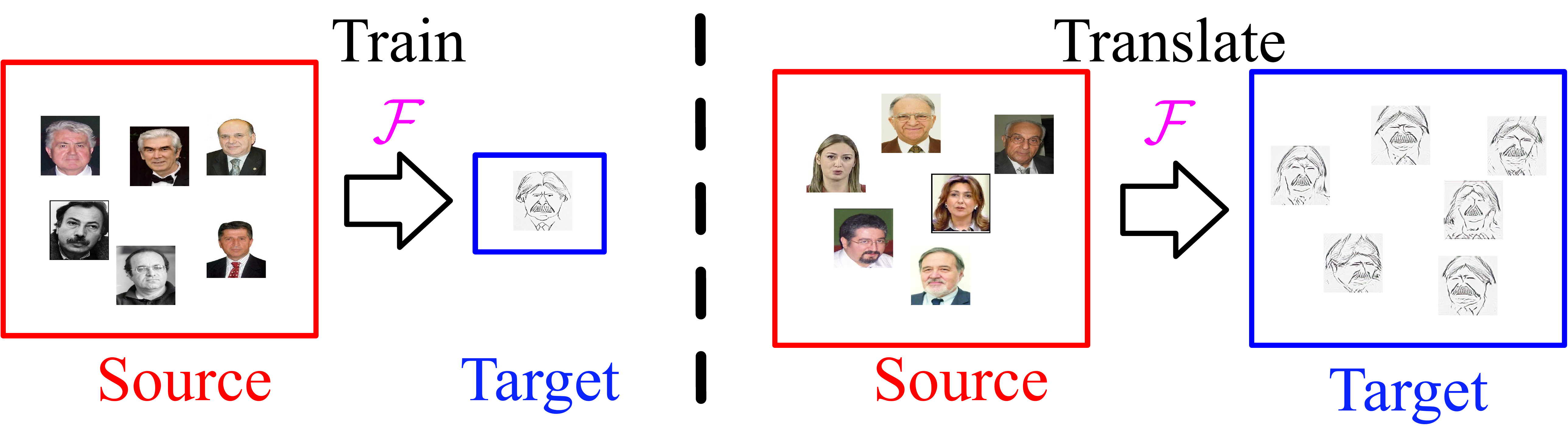}
\caption{One-shot cross-domain image-to-image translation. Note that we have only one target sample for training.}
\label{fig:demo}
\end{figure}

Benefited from the great success of deep learning based approaches, researchers have made much progress on computer vision fields such as image classification~\cite{shu2015weakly,wang2015singa,krizhevsky2012imagenet,he2016deep,simonyan2014very,szegedy2016rethinking}, image retrieval~\cite{potapov2018semantic,ji2017cross,wu2013online} and image recognition~\cite{he2015spatial,huang2017densely}. Generally, these methods may achieve reasonable results based on sufficient data for training the deep neural networks~\cite{krizhevsky2012imagenet,he2016deep}. However, collecting and labeling those data are time-expensive and tedious. In certain real-world scenarios, it may be impossible to gather abundant data from target domain $Y$ because of the scarcity of the image samples (In the worst case, it is possible to have only one image from $Y$). Nonetheless, we probably have redundant data from another source domain $X$, whose image samples are apparently correlated to the ones in the target domain $Y$ (such as photo and sketch images as shown in Fig.~\ref{fig:demo}). It would be a feasible solution if we generate images of domain $Y$ corresponding to analogous images of domain $X$ based on the diversity while keeping the semantic matching. Previous one-shot work mainly focuses on one-shot image recognition~\cite{inoue2018few, koch2015siamese,VinyalsBLKW16,duan2017one}. They try to find a meta-learning framework, which could easily adapt to a new task with slight fine-tuning on one sample. In this paper, we mainly concern about one-shot unpaired image-to-image translation problem. Our purpose is to find a mapping function $\mathcal{F}$ to translate images from the source image domain $X$ to the target image domain $Y$ with only one image sample as shown in Fig.~\ref{fig:demo}. Using image translation, we are able to enrich training samples of the target domain by translating images from a relevant source domain even if limited target domain samples are given. 

With regard to image-to-image translation field, Gayts et al.~\cite{Gatys2015A} first proposed Neural Style algorithm, which combines the content of one image with the style of another image using convolutional neural networks. Johnson et al. ~\cite{johnson2016perceptual} adopted the perceptual distance to measure the content and style similarity between different images. However, the translation result is confined to image painting style translation without high-level semantic matching. Advanced by the powerful ability of modelling visual content of generative adversarial networks (GANs), several recent research endeavours have been devoted to apply adversarial training to enhance the robustness and generality of traditional image-to-image translation~\cite{zhu2017unpaired,yi2017dualgan,huang2018multimodal,liu2017unsupervised,lee2018diverse,isola2016image,zheng2019generative}. These methods are able to obtain acceptable performance by using sufficient training data from both the target and the source domains. As aforementioned, we usually encounter the situation that the target domain does not have enough training samples. In certain cases, we only have one sample, which even has no counterpart in the source domain. 
\begin{figure}[!ht]
\centering
\includegraphics[width=\columnwidth]{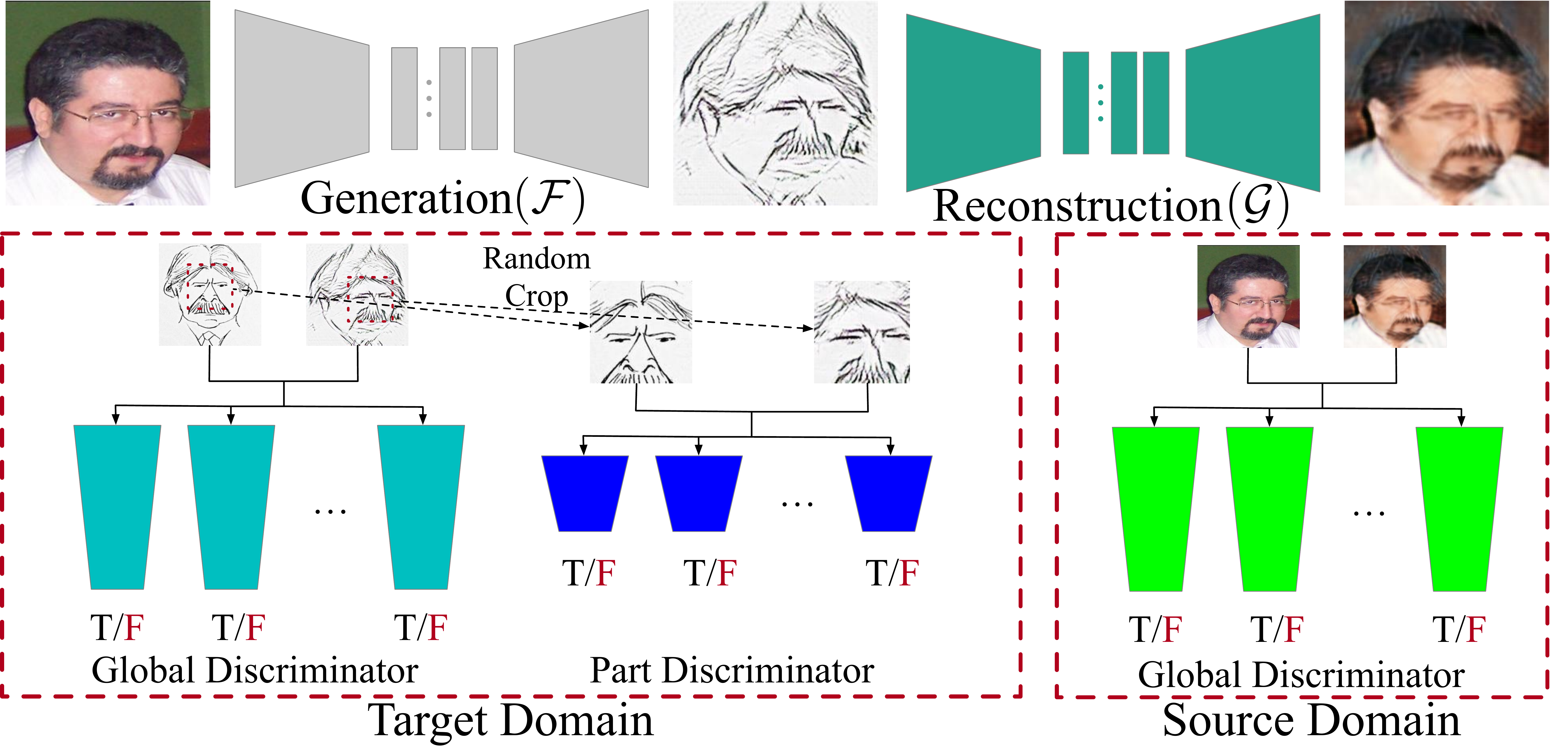}
\caption{The framework of our method. We take advantage of part-global learning together with a multi-adversarial discriminator architecture (MA). $\mathcal{F}$ aims to translate images from source domain to target domain while $\mathcal{G}$ tries to reconstruct the outputs of $\mathcal{F}$ in source domain.}
\label{fig:archi}
\end{figure}
The first attempt about one-shot unpaired cross-domain image-to-image translation~\cite{benaim2018one} focuses on one-to-many image-to-image translation scheme, which transforms the only one target sample into the source domain. Such assimilation process may easily cause the lost of the specific knowledge attached to the target sample. In contrast, we target at the many-to-one image-to-image translation, i.e., converting the diverse source samples to the target domain. We argue this is more challenging due of the extremely limited exploration of the target domain. To overcome the above obstacle, one feasible solution is to exploit the generative power of GANs to enable the many-to-one translation. Nevertheless, direct applying GANs may suffer from two major challenges: 1) the imbalance of insufficient target data and abundant source data leading to the overfitting during the learning process of the discriminator in the target domain; and 2) the lack of discriminative ability for extracting the high-level semantic representation, thereby failing to transfer the semantic information from the source domain to the target image domain.

Intuitively, learning from one sample in the human mind usually relies on part-to-part analogical reasoning in order to obtain fine-grained information. Inspired by such process, in this work, we devise a part-based discriminator, which is capable of distinguishing local part randomly cropped from translated images and real images using the limited information from the target domain. The benefits of designing the part-based discriminator are two-fold: 1) it helps to capture local characteristics of the target samples in a more accurate manner; and 2) it can assist the discriminator alleviating the overfitting problem by using random partial information rather than the entire image.

Besides, in order to balance the learning for the target data and the source data, we devise a balanced adversarial loss, which utilizes a controlling hyper-parameter to reduce the convergence speed of the objective function. It is worth noting that if we do not use this balanced adversarial loss, the model tends to suffer from the trivial solution and cause the overfitting problem, i.e., generating extremely similar sample to the only one target sample no matter what the input of the source domain is. Furthermore, following~\cite{zheng2019generative}, we divide a original discriminator into a bunch of weak learners via multiple threads, which not only helps to significantly improve the efficiency by reducing the number of the training parameters, but also digs in more fine-grained semantic details of the only one target sample.

The contribution of this paper is summarized as follows:

\begin{itemize}
\item We propose a novel and effective one-shot image-to-image translation framework to translate abundant images from a source domain to another target image domain containing only one image. To the best of our knowledge, our work is one of the first attempts to achieve the one-shot unpaired cross-domain image-to-image translation in many-to-one setting. 
\item We propose to utilize a multi-adversarial mechanism via part-global learning to enhance the ability of the discriminator in characterizing the fine-grained semantics as well as significantly improve the efficiency of the training process. 
\item We introduce a balanced adversarial loss function to alleviate the influence of the data imbalance between the target domain and the source domain.
\end{itemize}

The rest of this paper is organized as below. Section~\ref{sec:relatedwork} briefly introduces the related work and Section~\ref{sec:method} elaborates the proposed approach. Section~\ref{sec:experiments} presents the extensive experimental results on various datasets, followed by the conclusion in Section~\ref{sec:conclusion}.

\section{Related work}\label{sec:relatedwork}
\subsection{Image-to-image translation}
 Due to the success of conditional GAN~\cite{mirzaO14}, many popular image-to-image translation methods were developed such as Pix2pix~\cite{isola2016image} and Pix2pixHD~\cite{Chun2017}. They could achieve the high-resolution and precise translation by training with paired images. However, paired training data are not always available. To overcome this shortage, many unpaired image domain translation were proposed including CycleGAN~\cite{zhu2017unpaired}, DualGAN~\cite{yi2017dualgan}, DiscoGAN~\cite{kim2017learning}, UNIT~\cite{liu2017unsupervised}, MUNIT~\cite{huang2018multimodal}, DRIT~\cite{lee2018diverse}. These methods could translate images from one domain to another domain by using unpaired images. CycleGAN~\cite{zhu2017unpaired} adopted a cycle-consistent adversarial loss to constrain the reconstruction of target images. MUNIT~\cite{huang2018multimodal} used an unsupervised multimodal structure to translate styles as well as contents to reconstruct the target images. The concurrent DRIT~\cite{lee2018diverse} aimed to generate images with diverse outputs, which proposed a disentangled representation framework.  GANimorph~\cite{gokaslan2018improving} combined shape deformation based on a discriminator and dilated convolutions to perform cross-species translation. Besides, Twin-GAN~\cite{li2018twin} used a progressively growing skip connected encoder-generator structure for human-anime character translation. Nevertheless, most of these works mainly performed experiments with redundant images from both source and target domains, which could perform unsatisfactory when limited images are given.

\subsection{One-shot image translation}
One-shot learning, which was first discussed by Fei-Fei Li and Erik Miller~\cite{fei2002rapid,miller2000learning}, aims to learn information about object categories from one, or only a few, training samples. Most of published one-shot learning approaches focus on how to recognize objects from a few samples (one sample)~\cite{long2017learning,inoue2018few, zitian2018}.
Different from above one-shot object recognition methods, one-shot image translation (OST) aimed to translate images between two domains in which one domain only includes one or a few images. This concept was first discussed by Benaim et al.~\cite{benaim2018one}, who aimed to generate an analogous of $y$ in $X$, with a single image $y$ from domain $Y$ and a set of images from domain $X$. To find a mapping function, they shared some specific layers of one variational autoencoder~\cite{kingma2013auto} to add a strong constrain between domain translation. Unlike their task, we aim to perform a more challenging task, which discovers a semantic mapping function to translate a set of images from $X$ to $Y$, namely, we have reverse translation direction with OST methods~\cite{benaim2018one}. In our case, we aim to use the semantic link between domains and unearth perceptual similarity between $X$ and $Y$ with abundant images of $X$ and one sample image of $Y$ are given. 

\subsection{Multi-adversarial training}
Recently many methods have utilized multi-adversarial training mechanism to enhance generation performance, which ensemble different discriminators functionally. GMAN~\cite{durugkar2017generative} first adopted multiple discriminators for high quality image generation with fast and stable convergence. Multi-discriminator CycleGAN~\cite{hosseini2018multi}, which is an extension of CycleGAN, was proposed to enhance the speech domain adaption with a multiple discriminators architecture. MD-GAN~\cite{hardy2018md} was proposed to use a GAN with multiple discriminators on the distributed datasets. Most of studies have used multiple discriminators to give the generator with better guidance. Pix2pixHD~\cite{Chun2017} and MUNIT~\cite{huang2018multimodal} adopted multi-scale discriminator structure for high-resolution paired and multimodal unpaired image-to-image translation respectively. Recently GAN-MBD~\cite{zheng2019generative} proposed a multi-branch discriminator to reduce the parameter of discriminators and enhance the translation between species.

Based on the multi-adversarial training, the image generation and translation quality has made a comprehensive progress. For our purpose, with limited images given, we aim to use the multi-adversarial training mechanism to improve the image-to-image translation process and increase the possibility to establish a high-level semantic link between different domains.

\section{The Proposed Approach}\label{sec:method}
In this section, we elaborate the proposed approach for many-to-one image-to-image translation.

\subsection{Part-Global discriminators}
\begin{figure}[t]
\centering
\includegraphics[width=0.9\columnwidth]{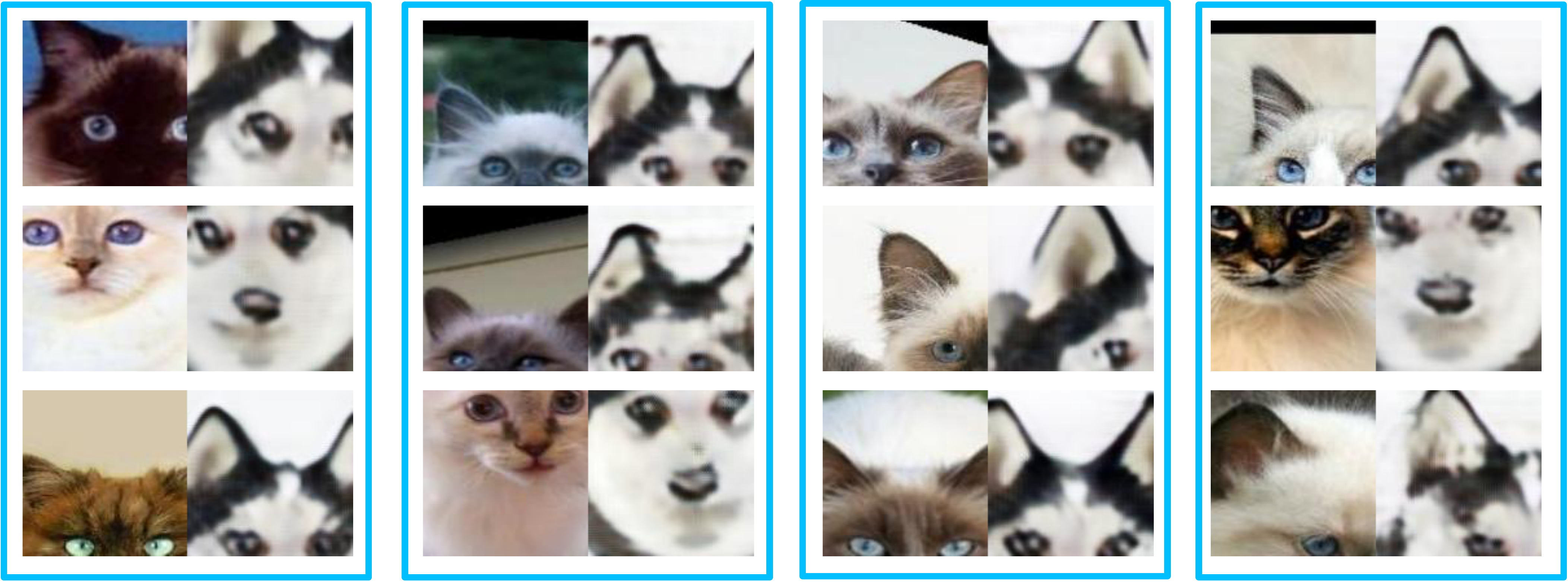}
\caption{Illustrative exemplars of translating ``cat'' to ``dog'' using proposed approach, which mimics analogical reasoning in the human mind while preserving global layout/pose information and achieving local semantic matching.}
\label{fig:part2part}
\end{figure}

As illustrated in Fig.~\ref{fig:part2part}, suppose we want to translate images from the source ``cat'' domain to the target ``dog'' domain with redundant ``cat'' samples and  only one ``dog'' sample, the intuitive principle of analogical reasoning is to 1) preserve global layout/pose of the original image, as well as 2) perform semantic matching of detailed parts, such as eyes, ears and nose.

Iizuka et al.~\cite{iizuka2017globally} proposed a global-local adversarial architecture to effectively combine global and local information to boost image inpainting. In particular, a local context discriminator was proposed to ensure local consistency, which makes sure the input of this local discriminator is a small area centered at the completed region. Inspired by the powerful modelling ability of local and global information, we devise a part-global adversarial architecture to increase the variety of the source domain and improve the one-shot image-to-image translation process. Specifically, our part discriminator is fed with a random part cropped from generated images and real images as shown in Fig.~\ref{fig:archi}. It is worth noting that our part discriminator is only designed for generator $\mathcal{F}$ in the target domain. Furthermore, we only feed a small part that randomly cropped from the entire real/fake image to enhance the robustness of models. By means of this method, more fine-grained part samples could be reachable by random cropping, thus our model could capture more detailed information from local context parts. To ensure the global consistency and semantic matching between generated images and the only one target sample image, we combine a common global discriminator to cope with the entire images. Note that we consider part discriminator and global discriminator as equal contribution in the learning process. The loss function can be described as:
\begin{equation}
\mathcal{L}(\mathcal{F},D)=
\begin{cases}
\mathcal{L}_{p}(\mathcal{F},D_{p}) \quad \text{for}\ D_{p},\\
\mathcal{L}_{g}(\mathcal{F},D_{g}) \quad \text{for}\ D_{g},
\end{cases}
\end{equation}
where 
\begin{equation}
\label{eq:gp}
\begin{split}
    \mathcal{L}_{g}(\mathcal{F},D_{g})=&\mathbb{E}_{y\sim p_{\text{data}}(Y)}[\log D_{g} (y)] +\\ &\mathbb{E}_{x\sim p_{\text{data}}(X)}[\log(1-D_{g}(\mathcal{F}(x)))],
\end{split}
\end{equation}
and 
\begin{equation}
\label{eq:gp1}
\begin{split}
    \mathcal{L}_{p}(\mathcal{F},D_{p})=&\mathbb{E}_{\hat{y}\sim p_{\text{data}}(Y)}[\log D_{p} (\hat{y})] +\\ &\mathbb{E}_{\hat{x}\sim p_{\text{data}}(X)}[\log(1-D_{g}(\mathcal{F}(\hat{x})))].
\end{split}
\end{equation}
Here $D_{p}$ and $D_{g}$ denote part discriminator and global discriminator respectively, $\mathcal{F}(\hat{x})$ and $\hat{y}$ denote the random part region cropped from generated images and real images respectively.

\subsection{Characterizing fine-grained semantics}
\begin{figure}[!ht]
\centering
\includegraphics[width=\columnwidth]{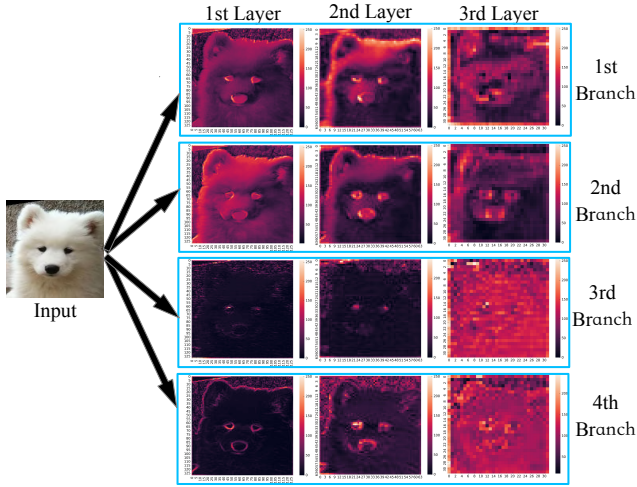}
\caption{The visualization of multi-adversarial discriminators for the only one target domain sample. Different threads capture different semantic details.}
\label{fig:visual}
\end{figure}

In order to characterize the detailed semantics in the image, we propose to utilize the ``divide-and-conquer'' strategy, i.e., designing different threads of discriminators to decide whether the current image and part region are real or synthesized. The formulation can be described as:
\begin{equation}
\begin{split}
    \mathcal{L}(\mathcal{F},D)=&\frac{1}{N}\sum_i^N{\{\mathbb{E}_{y\sim P_{\text{data}}(Y)}\left[\log D_i(y)\right]}\\
    &+\mathbb{E}_{x\sim P_{\text{data}}(X)}\left[\log (1-D_{i}(\mathcal{F}(x)) \right]\},
\end{split}
\end{equation}
where $N$ denotes the thread number of a common discriminator. We feed the average adversarial loss of discriminators to update generators, and each thread of discriminators is optimized independently. As described in ~\cite{zheng2019generative}, each thread of discriminators could learn a semantic sub-task automatically. With this implicit semantic division, our model could have stronger ability to unearth intrinsic link between source and target domains. As illustrated in Fig.~\ref{fig:visual}, to make an explicit elaboration, we train discriminators of 4 threads with abundant samples from source domain and the only one target sample, and visualize the feature map outputs of the target sample. Each thread of the model is able to take charge of a different semantic representation of the only one target sample. The first thread focuses on eyes and nose while the second thread pays attention to fur information. The third thread captures small detailed information, and the fourth thread observes the edge information. 

\subsection{Balance training between source and target}
In consideration of the possible extremely imbalance between the target and the source domains for \emph{one vs. many} case, the discriminators of the target domain could be easily over-fitted if we keep the same training speed between discriminators of the target and source domains. In order to alleviate this problem, we develop a balanced adversarial loss to slow down the convergence to the one-shot image. Here we design a strategy by using a hyper-parameters $\alpha$ to control the convergence speed of discriminators for two domains. For the mapping function $\mathcal{F}: X\rightarrow Y$ and $\mathcal{G}: Y\rightarrow X$, the balanced adversarial loss is defined as:
\begin{equation}
\label{eq5}
\mathcal{L} = \alpha \mathcal{L}_{(\mathcal{F},D_{Y})} + \mathcal{L}_{(\mathcal{G},D_{X})},
\end{equation}
where $D_{X}$ and $D_{Y}$ denote discriminators for source domain and target domain respectively. 

\section{Experiments}\label{sec:experiments}

\subsection{Datasets}
We evaluate our approach by comparing with the state-of-the-arts on six different datasets:\\
\textbf{Caricature}~\cite{akleman2000making} includes 200 paired caricature images, which deforms the facial feature of real images.\\
\textbf{IIIT-CFW}~\cite{mishra2016iiit} contains 1000 real image and 8928 annotated cartoon faces of famous characters of the world with varying profession of 100 public figures.\\
\textbf{CelebA+Portrait}~\cite{lee2018diverse} is a combined dataset derived from CelebA~\cite{liu2015faceattributes} and Wikiart\footnote{https://www.wikiart.org/}. In specific, 6453 images are selected from CelebA as the source domain, and 1814 images are selected from Wikiart as the target domain.\\
\textbf{Cat2dog} is a cropped image dataset including 871 cat images and 1364 dog images in total. We inherit this dataset from DRIT~\cite{lee2018diverse}, and we follow the same data split for training and testing.\\
\textbf{Day2night}~\cite{laffont2014transient} contains 100 paired day-night images in which 1+90 (\emph{one vs. many}) images are used for training.\\
\textbf{PHOTO-SKETCH}~\cite{zhang2011coupled,laffont2014transient} is a photo-to-sketch translation dataset, which contains paired facial photos and sketch images.

\subsection{Implementation details}
We mainly inherit the architecture from CycleGAN~\cite{zhu2017unpaired}. We extend the layers of discriminators with part-global discriminators to capture the high-level semantic representation and adopt the multi-adversarial training mechanism in our model. The part discriminators own fewer layers than global discriminators. To improve the generality and robustness of models, we use some common data augmentation approaches including flip, slightly rotation, and center crop. The hyper-parameter $\alpha$ mentioned in Eq.~\ref{eq5} is set as $0.1$ in all our experiments. We use Adam~\cite{kingma2014adam} to optimize our model and set learning rate as $0.0002$.

\subsection{Evaluation metrics}
To evaluate the effectiveness of different methods, we measure the translated quality by using the following criteria:\\
\textbf{Fr\'echet Inception Distance} (FID)~\cite{heusel2017gans} computes the similarity between the generated sample distribution and real data distribution. This method is a consistent and robust approach for evaluating the generated images~\cite{lucic2018gans,borji2019pros}, and it can be calculated by:
\begin{equation}
\text{FID}=||\mu_x-\mu_g||_2^2+Tr\left(\textstyle\sum_{x}+\sum_{g}-2(\sum_{x}\sum_{g})^{\frac{1}{2}}\right),
\end{equation}
where $(\mu_{x},\sum_{x})$ and $(\mu_{g},\sum_{g})$ are the mean and covariance of the sample embeddings from the data distribution and model distribution, respectively. Lower FID index means that the smaller distribution difference between the generated and the target images, and which represents higher generated image quality. In our one-shot unpaired image-to-image translation task, we can evaluate the image generation quality by computing these metrics. \\
\textbf{Learned Perceptual Image Patch Similarity} (LPIPS)~\cite{zhang2018unreasonable} computes the perceptual similarity between two images. A lower LPIPS means that the two images have more perceptual similarity. Considering two image domains, we can compute the LPIPS distance to evaluate the perceptual similarity.\\
\textbf{Structural Similarity} (SSIM)~\cite{hore2010image} is a traditional metric to measure the similarity between two images. Higher SSIM shows more structural similarity between generated images and real images.

\subsection{Comparison with the state-of-the-arts}
We compare our method to state-of-the-art image-to-image translation methods: CycleGAN~\cite{zhu2017unpaired}, MUNIT~\cite{huang2018multimodal} and DRIT~\cite{lee2018diverse}. The comparison is performed under two settings: \emph{one vs. many} and \emph{many vs. many}. For \emph{one vs. many} case, we use only one image from the target domain and many images from the source domain. For \emph{many vs. many} case, we use many images from both the two domains. We also compare our method with the OST~\cite{benaim2018one} method and fast-neural-style~\cite{johnson2016perceptual} method for \emph{one vs. many} case.

\subsubsection{Results on scene change}
\begin{figure}[!ht]
\centering
\includegraphics[width=\columnwidth]{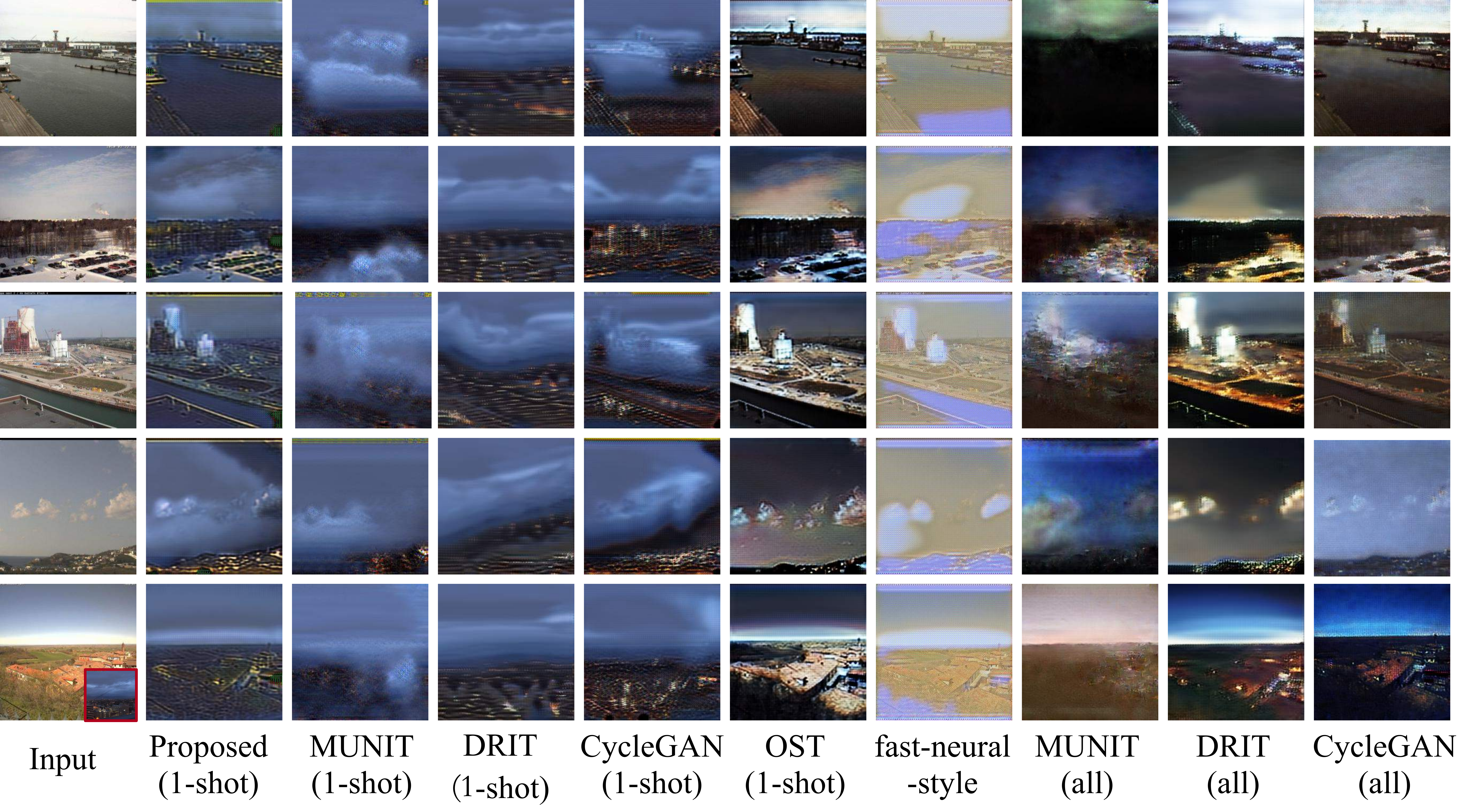}
\caption{The day$\rightarrow$night translation results on day2night dataset using different methods. The smaller image framed by red box in lower left corner input shows the only one training sample from target domain.}
\label{fig:day2night}
\end{figure}
We first conduct a scene change task on \textbf{day2night}~\cite{laffont2014transient} dataset. Fig.~\ref{fig:day2night} shows the translated results. It can be seen, our method generates satisfactory nighttime images from daytime inputs while using only one night style image as training sample for \emph{one vs. many} case. In contrast, DRIT generates images with several dirty color blocks, and fast-neural-style method merely achieves the color and textile translation, yielding unnatural synthesized images. The DRIT (1-shot) and MUNIT (1-shot) methods generate some detailed parts similar to the only one image from target domain without preserving some contents of input images. We compute the quantitative results (FID, LPIPS and SSIM) between outputs and paired ground truth images, and results are listed in Tab.~\ref{table:day2night_compare}. Our method has the lowest FID and highest SSIM value among all the methods.

\subsubsection{Results on photo-to-caricature}
In this part, we evaluate our method on a more challenging task, which aims to achieve the photo-to-caricature translation. Gats et al.~\cite{Gatys2015A} and Jonhson et al.~\cite{johnson2016perceptual} performed artist style transformation using one style image and a set of input images. We conduct experiments on four photo-to-caricature datasets, i.e., Caricature~\cite{akleman2000making}, PHOTO-SKETCH~\cite{zhang2011coupled,wang2009face}, IIIT-CFW~\cite{mishra2016iiit} and CelebA+Portrait~\cite{lee2018diverse}. This task requires not only satire exaggeration of photos but also artist style transfer. We compare our method to others under two settings: \emph{one vs. many} and \emph{many vs. many}.

\begin{table}[!ht]
\caption{Quantitative comparison of different methods for day$\rightarrow$night translation task on day2night dataset.}
\label{table:day2night_compare}
\begin{center}
\begin{scriptsize}
\begin{sc}
\begin{tabular}{lccc}
\toprule
Method & FID & LPIPS & SSIM\\
\midrule
Proposed (1-shot) & \textbf{223.3912} & 0.6887 & \textbf{0.6959}\\
DRIT (1-shot) & 290.4305  & 0.6913 & 0.6917 \\
MUNIT (1-shot) & 261.8025  & \textbf{0.6675} & 0.6636 \\
CycleGAN (1-shot) & 248.8642 & 0.6783 & 0.6823\\
OST (1-shot) & 355.2602 & 0.7063 & 0.5748\\
fast-neural-style  & 301.0206 & 0.7430 & 0.4672\\
\hdashline
DRIT (all) & 227.0346 & 0.6646 & 0.5555\\
MUNIT (all) & 225.1758 & 0.6654 & 0.5811\\
CycleGAN (all) & 327.0072 & 0.6885 & 0.7174\\
\bottomrule
\end{tabular}
\end{sc}
\end{scriptsize}
\end{center}
\end{table}

\begin{figure}[!ht]
\centering
\includegraphics[width=\columnwidth]{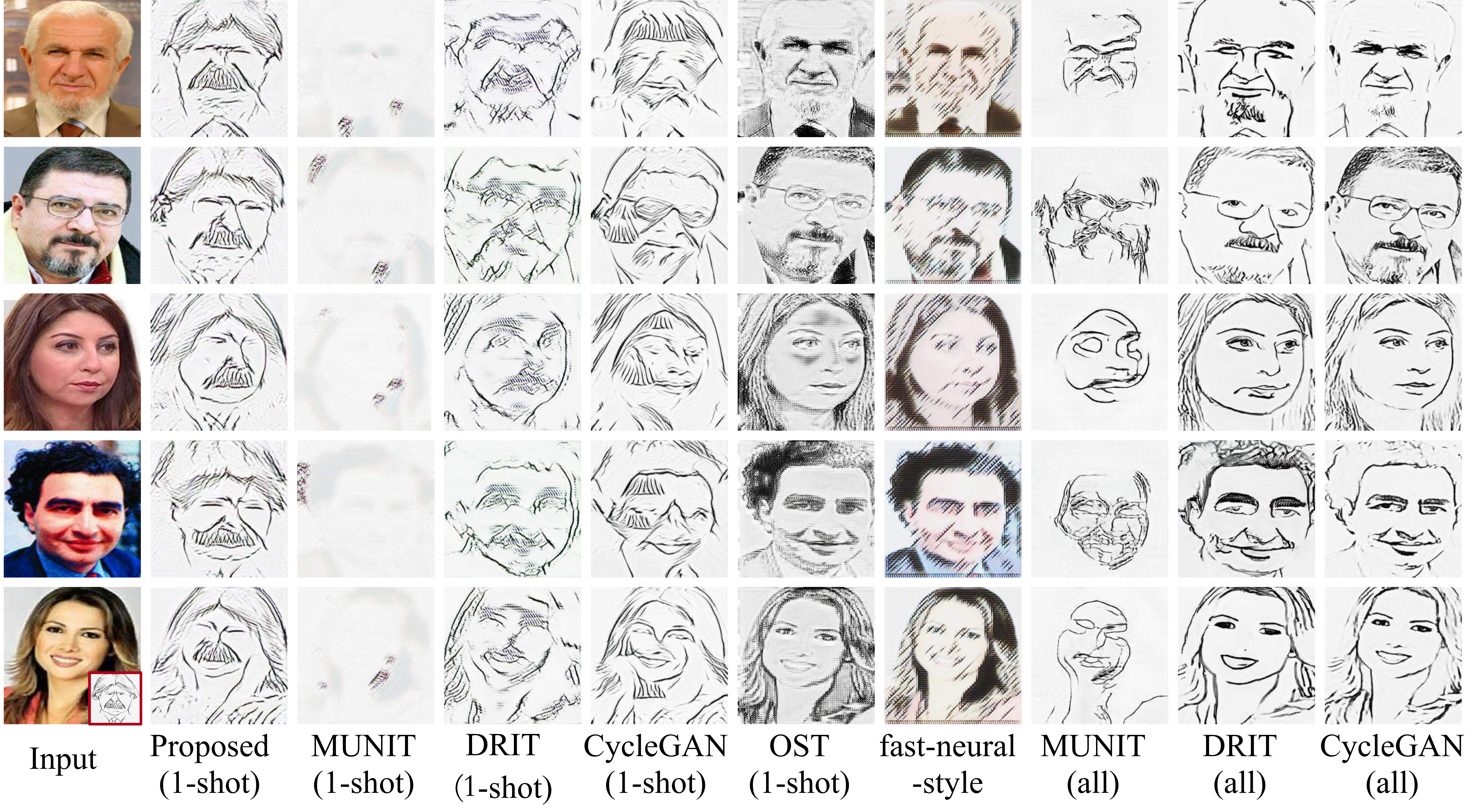}
\caption{The photo$\rightarrow$caricature translation results on Caricature dataset using different methods. The smaller image framed by red box in lower left corner input shows the only one training sample from target domain.}
\label{fig:cari_compare}
\end{figure}

\begin{table}[!ht]
\caption{Quantitative comparison of different methods for photo$\rightarrow$caricature translation task on Caricature dataset.}
\label{table:cari_compare}
\begin{center}
\begin{scriptsize}
\begin{sc}
\begin{tabular}{lccc}
\toprule
Method & FID & LPIPS & SSIM\\
\midrule
Proposed (1-shot) & \textbf{202.6208} & \textbf{0.5523} & \textbf{0.9629}\\
DRIT (1-shot) & 247.7226 & 0.5823 & 0.9577\\
MUNIT (1-shot) & 312.0169 & 0.7211 & 0.9616\\
CycleGAN (1-shot) & 272.2672 & 0.5813 & 0.9604\\
OST (1-shot) & 304.6888 & 0.6762 & 0.9076\\
fast-neural-style  & 278.7496 & 0.6507 & 0.8554\\
\hdashline
DRIT (all) & 97.8156 & 0.5506 & 0.9491\\
MUNIT (all) & 144.7875 & 0.6019 & 0.9552 \\
CycleGAN (all) & 133.4060 & 0.5445 & 0.9547\\
\bottomrule
\end{tabular}
\end{sc}
\end{scriptsize}
\end{center}
\end{table}

Fig.~\ref{fig:cari_compare} reports the translation results using different methods on Caricature dataset. The \emph{one vs. many} task used only one randomly-selected caricature image as the target sample and 160 photos, while the \emph{many vs. many} tasks used 160 photo images and 160 caricature images. The rest 40 pairs were used for testing. For the \emph{one vs. many} case, our method not only captures the caricature style but also preserves the pose, the layout and identity information of inputs. As can be observed from the second column of Fig.~\ref{fig:cari_compare}, our method generates an exaggerated beard on appropriate part while CycleGAN only generates a beard artifact on the same part region. DRIT generates images with blur boundary and some artifacts while MUNIT fails to synthesize satisfactory results. OST and fast-neural-style methods only obtain colored outputs according to the only one target sample without caricature translation. Tab.~\ref{table:cari_compare} presents the FID, LPIPS and SSIM values between generated images and ground truth images for the evaluated methods. Our method achieves the lowest LPIPS distance and highest SSIM score compared to the other methods.

\begin{figure}[!ht]
\centering
\includegraphics[width=\columnwidth]{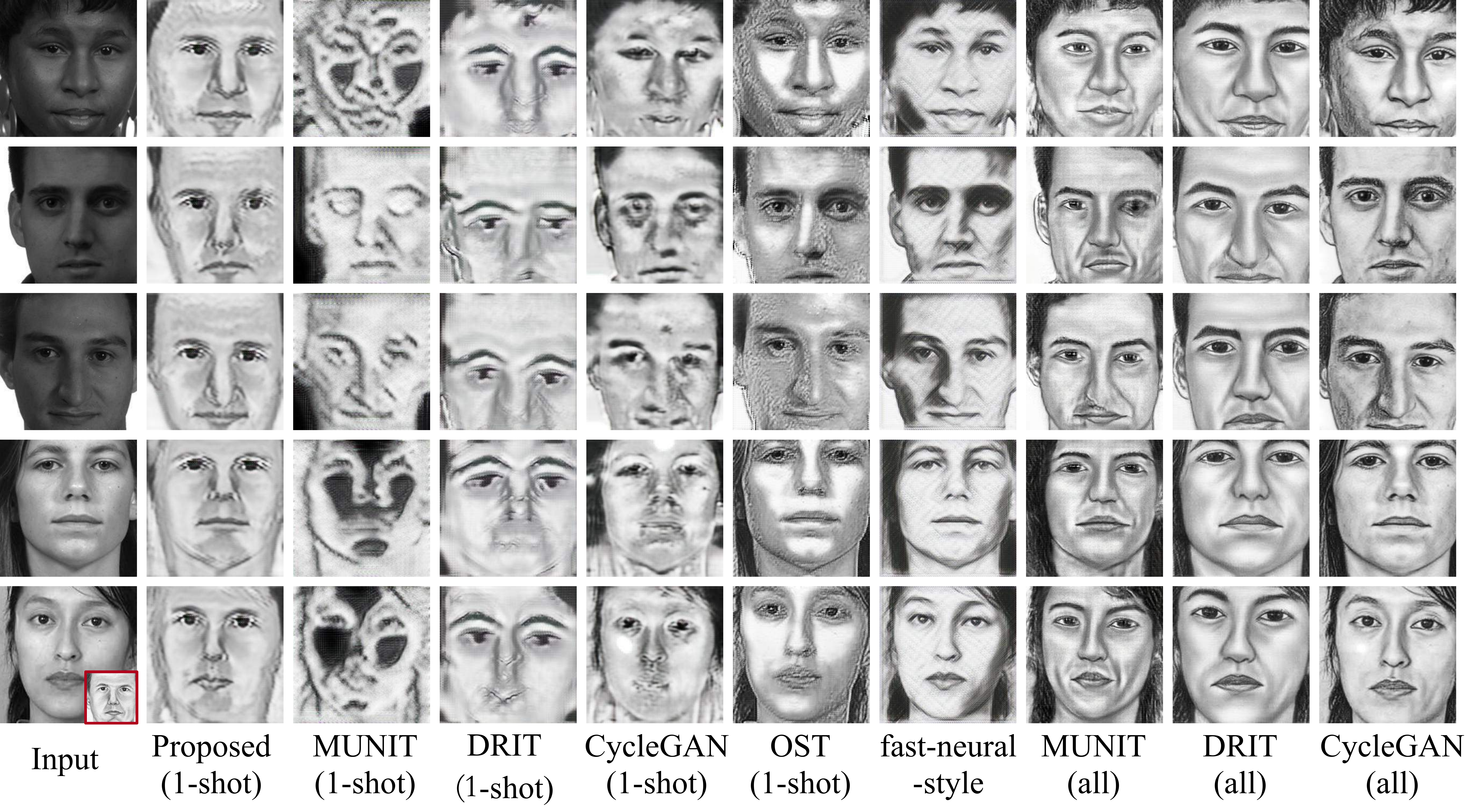}
\caption{The photo$\rightarrow$sketch translation results on PHOTO-SKETCH dataset using different methods. The smaller image framed by red box in lower left corner input shows the only one training sample from target domain.}
\label{fig:sketch_compare}
\end{figure}

\begin{table}[!ht]
\caption{Quantitative comparison of different methods for photo$\rightarrow$sketch translation task on PHOTO-SKETCH dataset.}
\label{table:sketch_compare}
\begin{center}
\begin{scriptsize}
\begin{sc}
\begin{tabular}{lccc}
\toprule
Method & FID & LPIPS & SSIM\\
\midrule
Proposed (1-shot) & \textbf{94.2810} & 0.3995 & \textbf{0.9172}\\
DRIT (1-shot) & 115.7270 & 0.5400 & 0.9037\\
MUNIT (1-shot) & 370.2599 & 0.5452 & 0.8127\\
CycleGAN (1-shot) & 192.3583 & 0.5941 & 0.9086\\
OST (1-shot) & 141.8420 & 0.4846 & 0.9140\\
fast-neural-style  & 143.1430 & \textbf{0.3603} & 0.9076\\
\hdashline
DRIT (all)  & 25.9333 & 0.2418 & 0.9554\\
MUNIT (all) & 41.1899 & 0.2782 & 0.9392 \\
CycleGAN (all) & 34.5234 & 0.2478 & 0.9471\\
\bottomrule
\end{tabular}
\end{sc}
\end{scriptsize}
\end{center}
\end{table}

Fig.~\ref{fig:sketch_compare} illustrates the translated results of different approaches on PHOTO-SKETCH~\cite{zhang2011coupled,wang2009face} dataset, which has consistent caricature style. Tab.~\ref{table:sketch_compare} shows quantitative performance of different methods. We used 995 photos and one randomly selected sketch image for the source domain and the target domain, respectively. The rest 199 photo-sketch image pairs are used for testing. As seen, all the methods can get plausible results when using all 995 training paired images. Nonetheless, when only one target sample is fed, most of them achieve poor performance. In contrast, our method, which gains the lowest LPIPS, the lowest FID and the highest SSIM performance, is able to well preserve the pose/layout information of the source samples and generate vivid sketch similar to the only one target style.

\begin{figure}[!ht]
\centering
\includegraphics[width=\columnwidth]{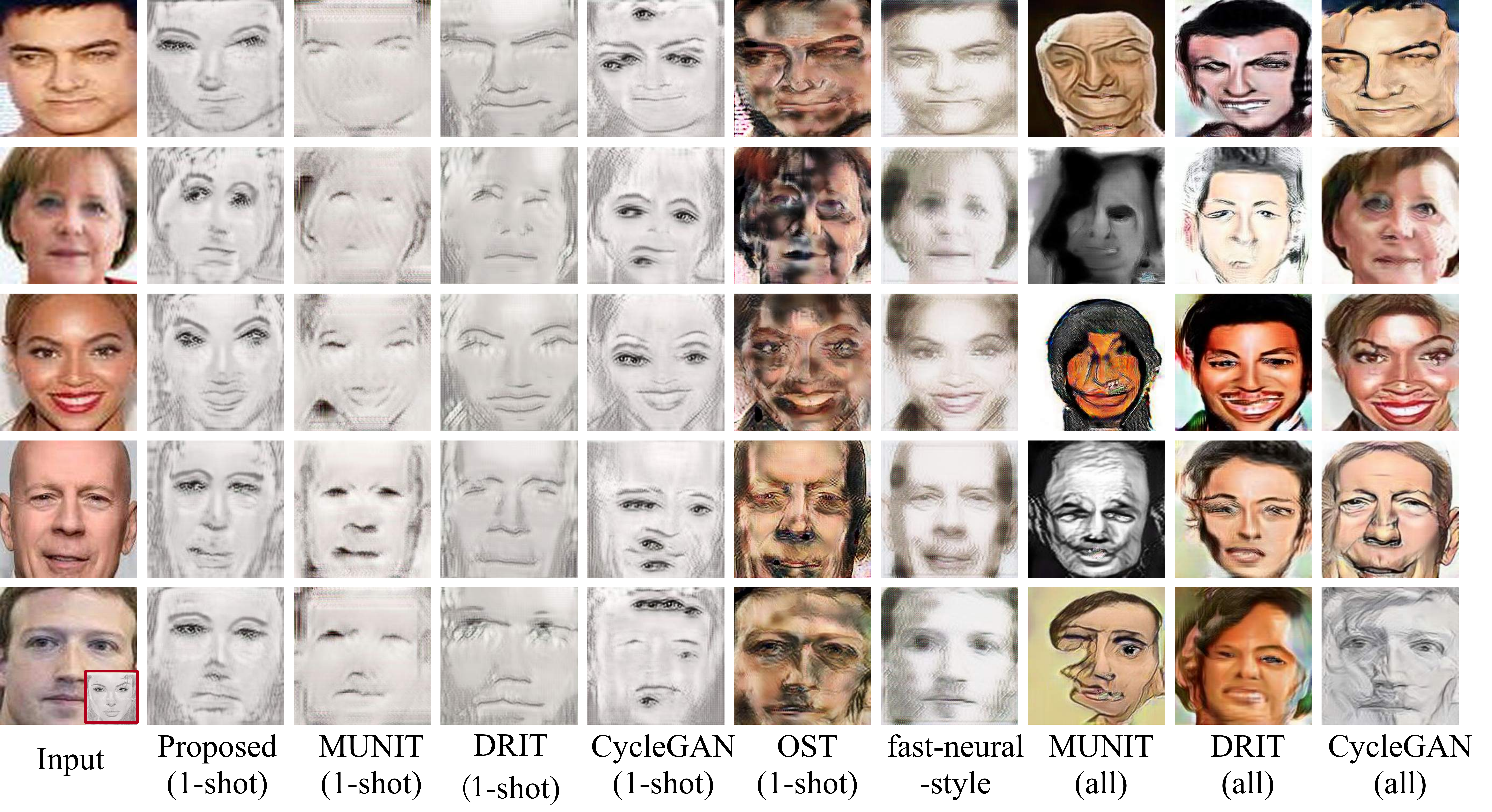}
\caption{The photo$\rightarrow$caricature translation results on IIIT-CFW dataset using different methods. The smaller image framed by red box in lower left corner input shows the only one training sample from target domain.}
\label{fig:IIIT_compare}
\end{figure}

Further, we conducted experiments on IIIT-CFW~\cite{mishra2016iiit} dataset. We randomly selected one image from caricature domain as training target and 800 photo images as training source. For \emph{many vs. many} case, we used all the caricature images as target. The rest 200 photo images are used for testing. Since IIIT-CFW does not provide real-cartoon image pairs, we only report the FID performance between the synthesized images and the real images as illustrated in Tab.~\ref{table:iiit_portrait}. Our approach method achieves the best results among all the evaluated methods in \emph{one vs. many} case. As illustrated in Fig.~\ref{fig:IIIT_compare}, compared to other methods under \emph{one vs. many} that merely perform textile transformation, our method can better characterize the semantic aspects (e.g., eyes and eyebrows), preserve the layout/pose information from source inputs, as well as inherit the style from the only one target sample.

\begin{figure}[!ht]
\centering
\includegraphics[width=\columnwidth]{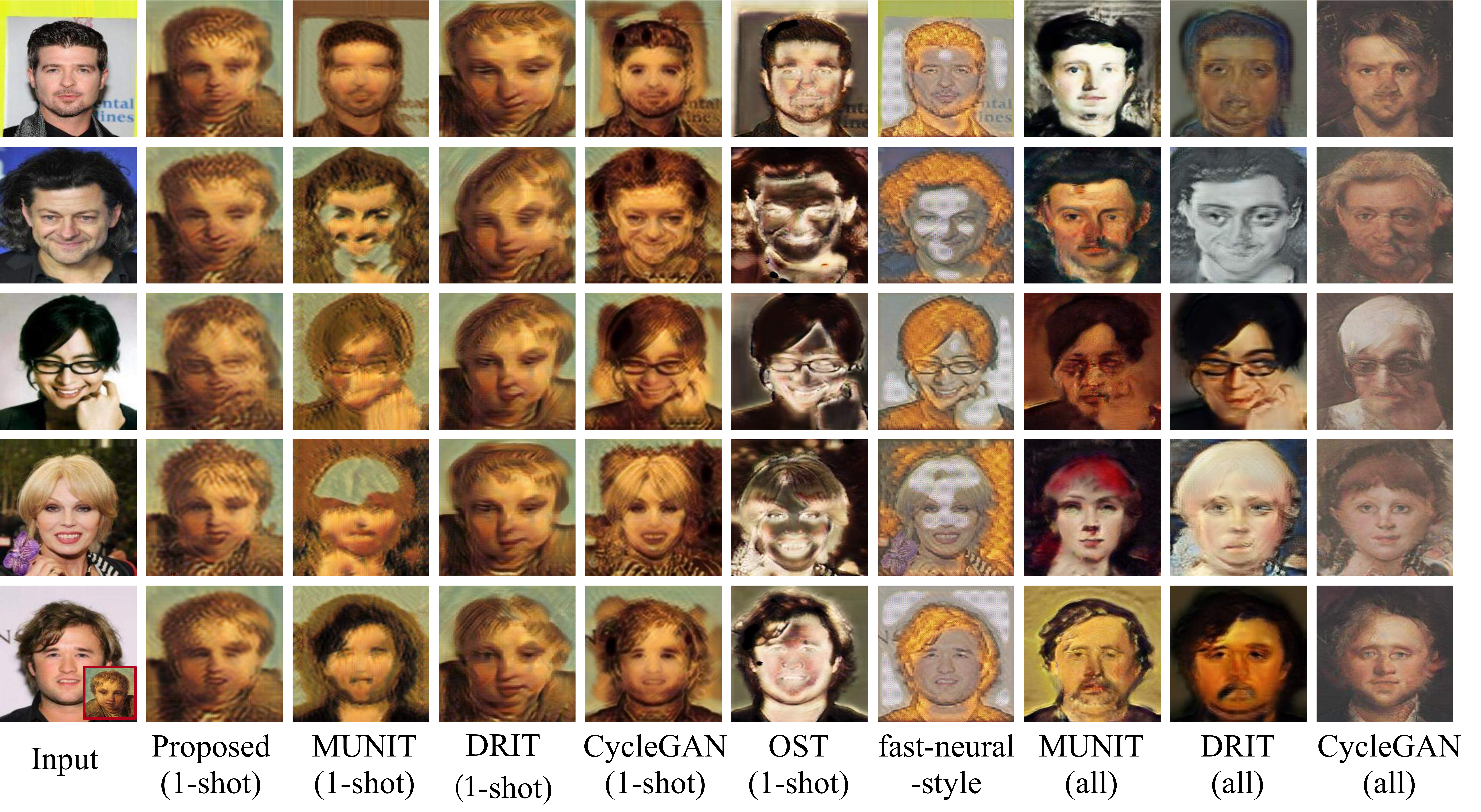}
\caption{The photo$\rightarrow$portrait translation results on CelebA+Portrait using different methods. The smaller image framed by red box in lower left corner input shows the only one training sample from target domain.}
\label{fig:portrait_compare}
\end{figure}

We also perform portrait translation using CelebA+Portrait~\cite{lee2018diverse} dataset. We follow the training/testing setting as in~\cite{lee2018diverse}. Since the source photos and the portrait images are not paired, we only compute the FID performance. We report the quantitative comparison results in Tab.~\ref{table:iiit_portrait}, from which we can observe that our proposed approach outperforms other compeitiors. Furthermore, as illustrated in Fig.~\ref{fig:portrait_compare}, in \emph{one vs. many} case, the compared methods either encounter overfitting problem with almost the same outputs for all the source inputs (i.e., DRIT), or achieve unacceptable traslation results (i.e., MUNIT). In contrast, our method can effectively preserve semantic details of source inputs (e.g., accessories, glasses), as well as the global layout/pose knowledge.

\begin{table}[!ht]
\caption{FID scores of different methods on photo$\rightarrow$caricature translation task on IIIT-CFW and CelebA+Portrait datasets. Smaller FID scores show better translation quality between outputs and real images from target domain.}
\label{table:iiit_portrait}
\begin{center}
\begin{scriptsize}
\begin{sc}
\begin{tabular}{lcc}
\toprule
Method & IIIT-CFW & CelebA+Portrait\\
\midrule
Proposed (1-shot) & \textbf{144.7352} & \textbf{144.5271} \\
DRIT (1-shot) & 178.5234& 183.6345\\
MUNIT (1-shot) & 263.3436& 248.2363\\
CycleGAN (1-shot) & 178.3645& 156.6452\\
OST (1-shot) & 245.1676 & 179.4688\\
fast-neural-style  & 166.3725 & 278.7496\\
\hdashline
DRIT (all) & 121.2450 & 139.1502\\
MUNIT (all) & 139.609 & 130.5828\\
CycleGAN (all) & 115.5868 & 131.4535\\
\bottomrule
\end{tabular}
\end{sc}
\end{scriptsize}
\end{center}
\end{table}

\begin{figure}[!ht]
\centering
\includegraphics[width=\columnwidth]{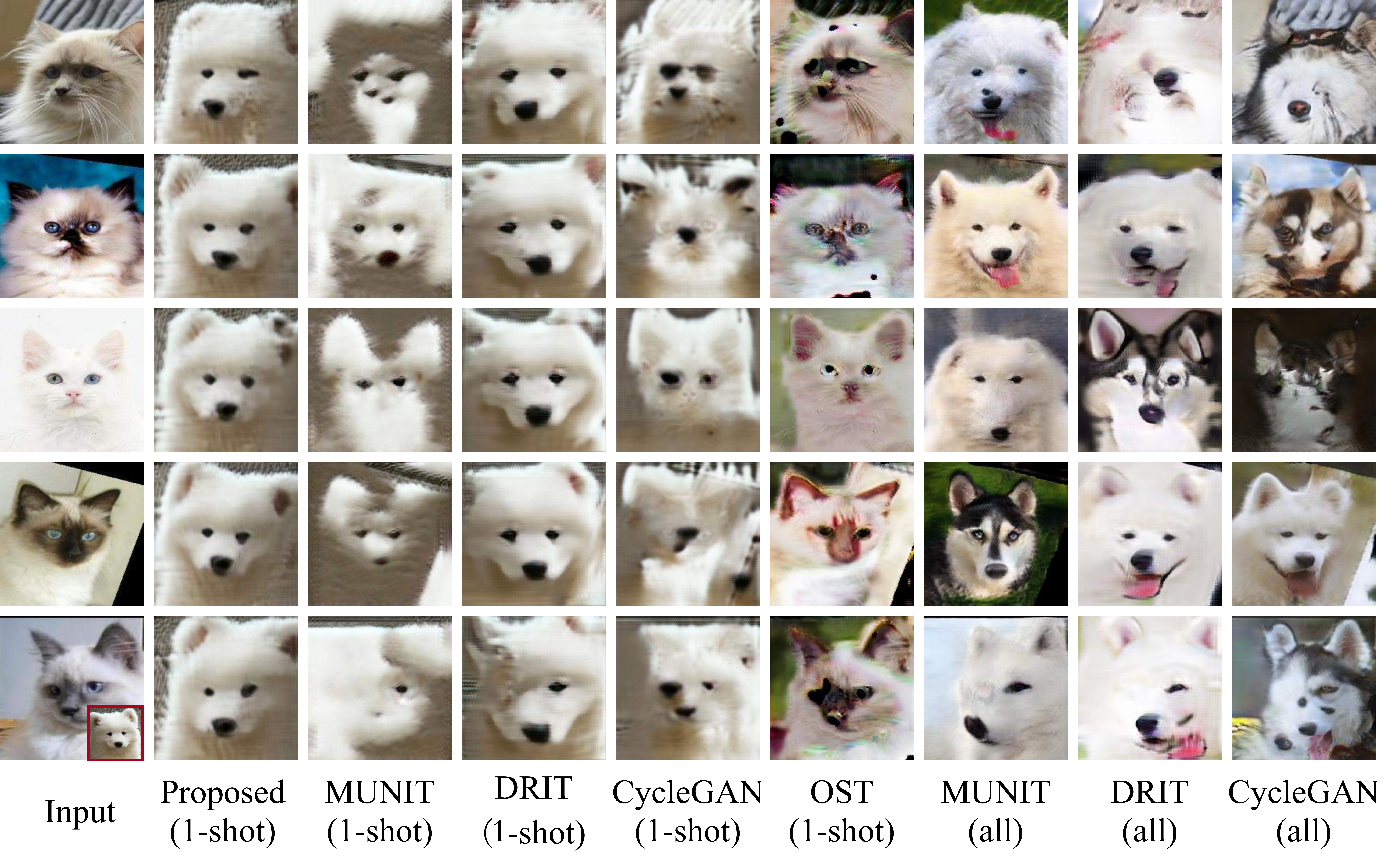}
\caption{The cat$\rightarrow$dog translation results on cat2dog dataset using different methods. The smaller image framed by red box in lower left corner input shows the only one dog image from target domain.}
\label{fig:cat2dog}
\end{figure}

\begin{figure}[!ht]
\centering
\includegraphics[width=\columnwidth]{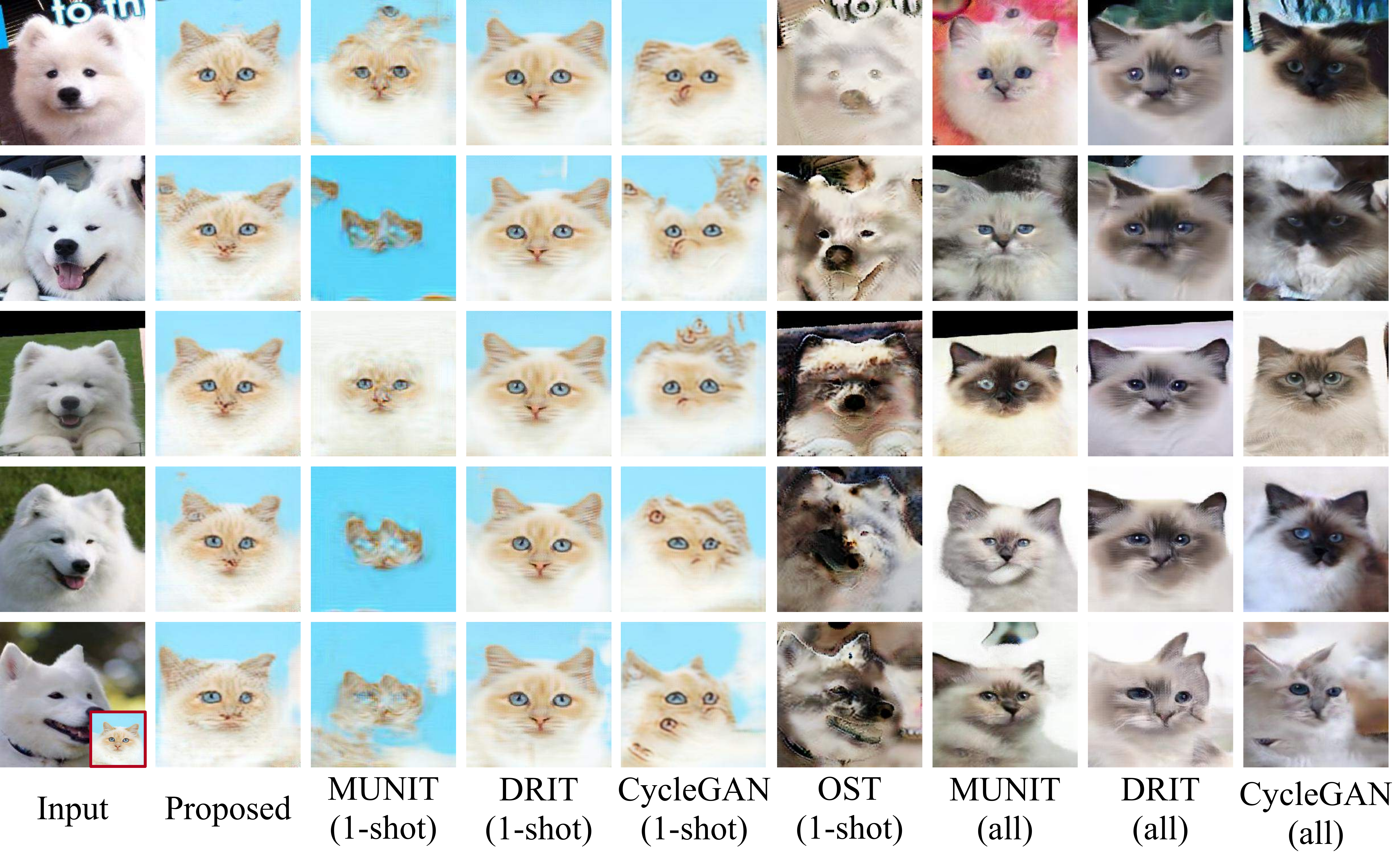}
\caption{The dog$\rightarrow$cat translation results on cat2dog dataset using different methods. The smaller image framed by red box in lower left corner input shows the only one cat image from target domain.}
\label{fig:dog2cat}
\end{figure}

\begin{table}[!ht]
\caption{FID scores of different methods for cat$\leftrightarrow$dog translation tasks on cat2dog dataset.}
\label{table:cat2dog_compare}
\begin{center}
\begin{scriptsize}
\begin{sc}
\begin{tabular}{lcc}
\toprule
Method & dog to cat & cat to dog\\
\midrule
Proposed (1-shot) & \textbf{124.7547} & \textbf{66.6420} \\
DRIT (1-shot) & 127.6726  & 123.6576 \\
MUNIT (1-shot) & 212.2386  & 222.1404 \\
CycleGAN (1-shot) & 155.1251 & 261.5311\\
OST (1-shot) & 277.0846 & 323.9576\\
\hdashline
DRIT (all) & 58.4392 & 88.6275\\
MUNIT (all) & 46.2864 & 47.4142\\
CycleGAN (all) & 46.9727 & 64.3956\\
\bottomrule
\end{tabular}
\end{sc}
\end{scriptsize}
\end{center}
\end{table}

\subsubsection{Results on cat$\leftrightarrow$dog}
We also evaluate our model on cat2dog dataset, which is a more challenging task to perform cross-species image-to-image translation. Fig.~\ref{fig:cat2dog} shows the cat$\rightarrow$dog translation results using different methods (only one dog image). Our method can preserve the layout/pose information and achieve the feature matching in high level space. OST method fails to achieve cross-species semantic translation. The MUNIT (1-shot) method achieves unsatisfactory results and generates imprecise semantic features, such as nose and eyes. Compared to DRIT (1-shot), our method can obtain better results and more detailed information. Contrary to Fig.~\ref{fig:cat2dog}, Fig.~\ref{fig:dog2cat} shows the dog$\rightarrow$cat translation results (only one cat image). The MUNIT (1-shot) method is unable to achieve precise translation while DRIT (1-shot) method may lead to overfitting problem, thereby generating similar images with the one training image. Compared to the methods trained using all images, our method cannot generate various background. Tab.~\ref{table:cat2dog_compare} shows the FID results using all methods. Our method obtains the best performance compared to the other methods for \emph{one vs. many} case. 

\subsection{Ablation study}
\label{sec:ablation}
\begin{figure}[!ht]
\centering
\includegraphics[width=\columnwidth]{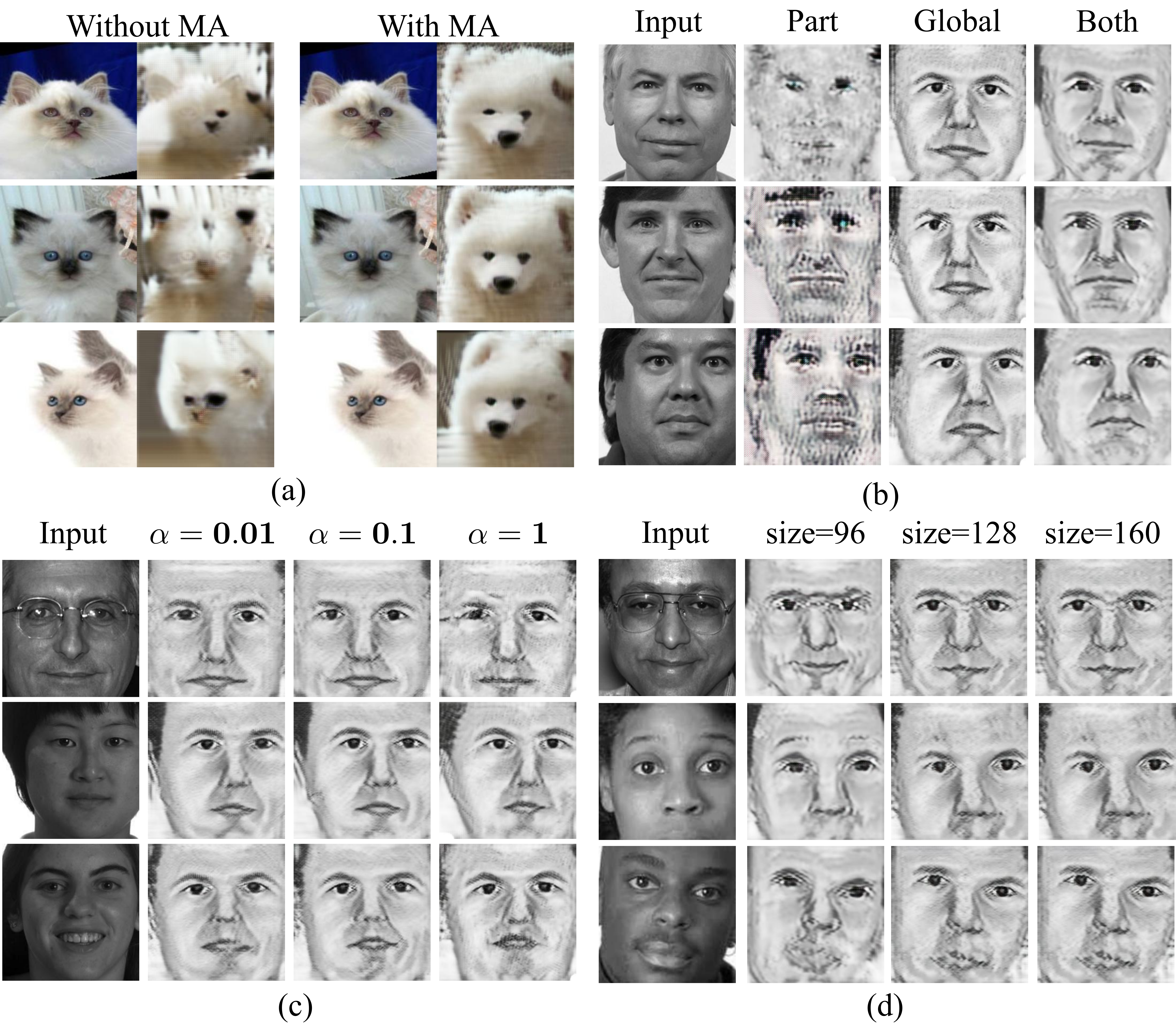}
\caption{The effectiveness of (a) multi-adversarial discriminators, (b) part discriminator and global discriminator, (c) the hyper-parameter $\alpha$ and (d) the size of cropped region.}
\label{fig:ablation}
\end{figure}

To investigate the efficacy of different components in our approach, we design several additional experiments for ablation study. Tab.~\ref{table:ablation_compare} shows the FID values of different variants of our approach on cat$\leftrightarrow$dog task. As can be seen, the performance drops dramatically by comparing GAN only with GAN+MA, which indicates that when using the multi-adversarial discriminators, the model can increase the ability to capture fine-grained semantic information. Fig.~\ref{fig:ablation}(a) also gives the visualized results of the similar conclusion. Fig.~\ref{fig:part2part} exhibits image translation results with local semantic matching while using additional part discriminators. We note that the translated images have a corresponding semantic link (mainly pose and layout) to their inputs. 

The effectiveness of global discriminators and part discriminators can be shown in Fig.~\ref{fig:ablation}(b) and Tab.~\ref{table:ablation_compare}. It is easy to find that the translated images have very poor global consistency without global constraints if we do not use global discriminators. When we only use global discriminators, the generated images look similar to the same artifacts on the same region in all generated images. When using part discriminators, the performance can be apparently improved, which benefits from the fact that the part discriminator can help alleviate the over-fitting by enforcing the generator to pay attention to the more fine-grained semantic details. 

Additionally, we investigate the effects of the part size of part discriminators and hyper-parameter $\alpha$. Tab.~\ref{table:part_compare} shows the experimental results on PHOTO-SKETCH dataset and Fig.~\ref{fig:ablation}(c) exhibits results using different part sizes. Fig.~\ref{fig:ablation}(d) shows some examples by using different values of $\alpha$. We get the best performance when we choose the part size as 128 and $\alpha=0.1$. 

\begin{table}[!ht]
\caption{FID scores of ablation study for cat$\leftrightarrow$dog translation tasks on cat2dog dataset.}
\label{table:ablation_compare}
\begin{center}
\begin{scriptsize}
\begin{sc}
\begin{tabular}{lcc}
\toprule
Method & dog$\rightarrow$cat & cat$\rightarrow$dog\\
\midrule
GAN only & 398.6242  & 413.3073 \\
GAN + Balance & 336.3632  & 285.0524 \\
GAN + MA & 146.3534 & 90.8875\\
GAN + Part + MA & 189.6234 & 122.5243\\
GAN + Global + MA & 138.6345 & 84.2634\\
GAN + Part + Global + MA & 132.3326 & 73.2645\\
Proposed & \textbf{124.7547} & \textbf{66.6420} \\
\bottomrule
\end{tabular}
\end{sc}
\end{scriptsize}
\end{center}
\end{table}

\begin{table}[!ht]
\caption{Quantitative results of using different part sizes and values of $\alpha$ for photo$\rightarrow$sketch translation task on PHOTO-SKETCH dataset.}
\label{table:part_compare}
\begin{center}
\begin{scriptsize}
\begin{sc}
\begin{tabular}{lccc}
\toprule
Method & FID & LPIPS & SSIM\\
\midrule
96 (part size) & 116.2623 & 0.4273 & 0.9026\\
128 (part size) & \textbf{94.2810} & \textbf{0.3995} & \textbf{0.9172}\\
160 (part size) & 105.0279 & 0.4078 & 0.9030\\
\hdashline
0.01 ($\alpha$) & 116.3350 & 0.4042 & 0.8961\\
0.1 ($\alpha$) & \textbf{94.2810} & \textbf{0.3995} & \textbf{0.9172}\\
1.0 ($\alpha$) & 137.5862 & 0.4668 & 0.8953\\
\bottomrule
\end{tabular}
\end{sc}
\end{scriptsize}
\end{center}
\end{table}

\subsection{Limitation and failure cases}
\begin{figure}[!ht]
\centering
\includegraphics[width=0.8\columnwidth]{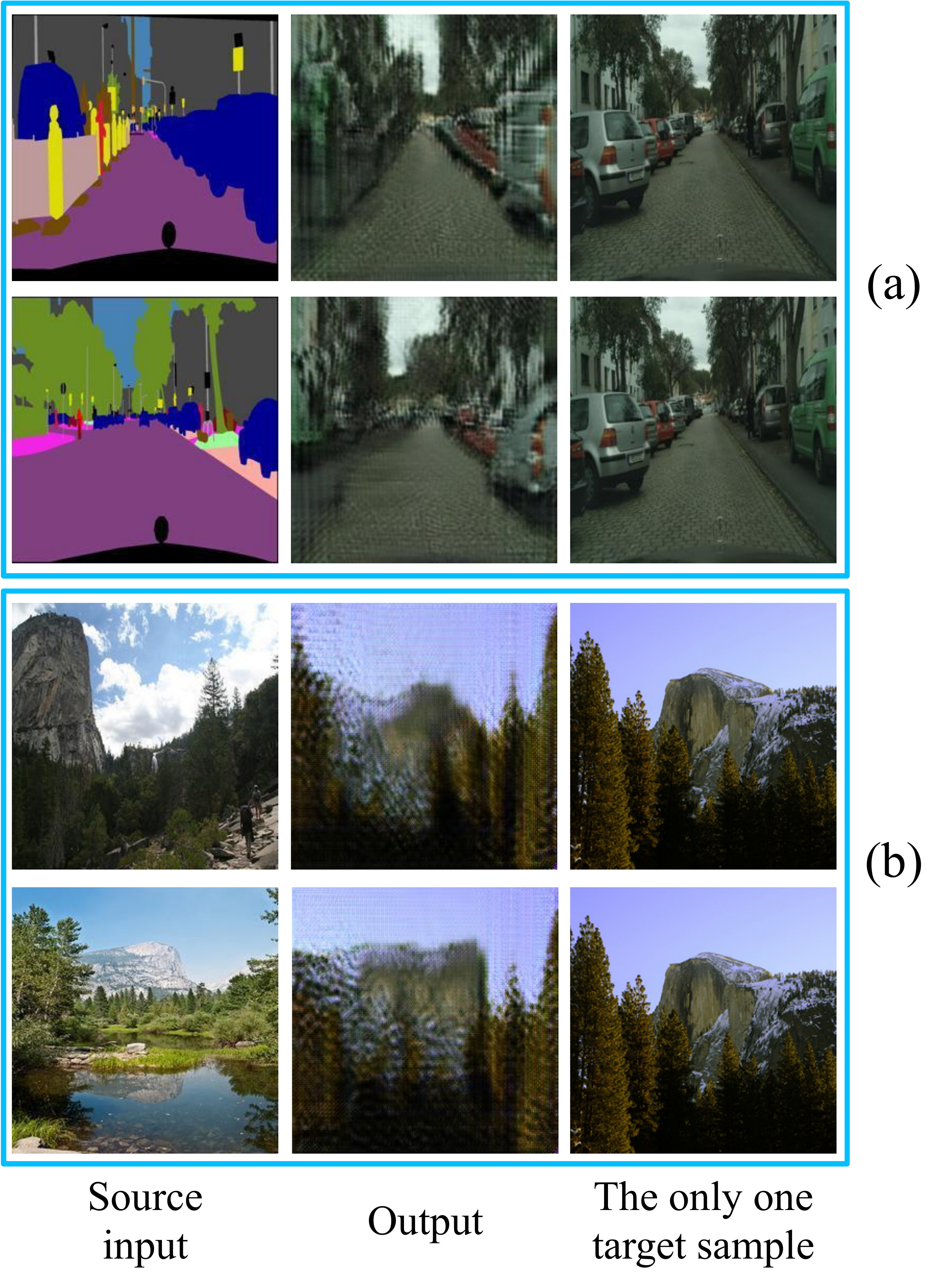}
\caption{Several failure cases of our method on Cityscapes and summer$\rightarrow$winter datasets.}
\label{fig:failure}
\end{figure}

We evaluated our method on a variety of one-shot image-to-image translation tasks, and the results were not always satisfactory. We analyze the underlying reason causing such phenomenon is the limitation of our approach for handling ``unknown'' objects. During the translation process, if the source object and the target object are not apparently correlated, then the translation becomes difficult. For instance, in the cat$\leftrightarrow$dog task, though we can accomplish the translation of the main object from cat to dog, the background of the source image cannot be preserved. One possible reason is that ``cat'' and ``dog'' are semantically similar to each other, while the background is not necessarily correlated to ``dog'', thereby leading to the ``abundance'' of the background in the output image. We show more experimental results of failure cases on Cityscapes~\cite{Cordts2016Cityscapes} and summer$\rightarrow$winter~\cite{zhu2017unpaired}. Fig.~\ref{fig:failure}(a) shows the failure results of Cityscapes dataset. When the scene is complex and the only one image from target domain could not cover all semantic information in target domain on this semantic generation task. In Fig.~\ref{fig:failure}(b), we conduct summer-to-winter translation while using one image depicting winter scene. As can be observed, the translation fails to preserve ``lake'' in the generated image and the ``cloud'' is translated to ``mountain'' by mistake. Such phenomenons indicate that our approach tends to fail if the objects (e.g., ``cloud'', ``lake'') have never been observed in the target domain.

\section{Conclusion and future work}\label{sec:conclusion}
In this paper, we proposed an effective method for one-shot cross-domain image-to-image translation to translate abundant samples from a source domain to another target domain with only one image. We introduced multi-adversarial scheme to enhance the ability of discriminators to unearth effective information with given limited images. Besides, we included a part-global learning architecture to extract more fine-grained information. Last but not least, we present a balanced adversarial loss to stabilize the adversarial training process and avoid over-fitting. We validated our method on multiple datasets and proved that our model is able to make use of the diversity information from the source domain and generate various kinds of images for the target domain even if the target domain only contains one training sample.

In future, we intend to dig in the interpret-ability of our model, which can enforce additional conditional information on the model to focus on semantic representation of images. 

\bibliographystyle{ACM-Reference-Format}
\bibliography{sample-base}
\end{document}